\documentclass[runningheads]{llncs}

\usepackage[T1]{fontenc}
\usepackage{graphicx}
\usepackage{booktabs}
\usepackage{url}
\usepackage{multirow}
\usepackage{amsmath}
\usepackage[colorlinks=false,hidelinks]{hyperref}
\usepackage{color}

\usepackage{bbding}

\titlerunning{BLUEX v2: Benchmarking LLMs on Open-Ended Questions}

\begin{document}

\tolerance=999
\sloppy

\title{BLUEX v2: Benchmarking LLMs on Open-Ended Questions from Brazilian University Entrance Exams}

\author{Jo\~ao Guilherme Alves Santos\inst{1,2}\orcidID{0000-0001-5307-5338}\textsuperscript{(\Envelope)} \and
Giovana Kerche Bonás\inst{1,2,3}\orcidID{0009-0001-9460-8353} \and 
Thiago Laitz\inst{1,2,3}\orcidID{0000-0001-7205-2094} \and 
Thales Sales Almeida\inst{1,2,3}\orcidID{0009-0006-9568-9331} \and
Helio Pedrini\inst{1}\orcidID{0000-0003-0125-630X}}

\authorrunning{Santos et al.}

\institute{University of Campinas (UNICAMP), Campinas-SP, Brazil \and
Tropic AI, Campinas-SP, Brazil  \and
Maritaca AI, Campinas-SP, Brazil \\
\email{j199624@dac.unicamp.br}}
\maketitle
%
\newcommand{\repourl}{\url{https://github.com/TropicAI-Research/BLUEXv2}}

\begin{abstract}
Although Large Language Models (LLMs) excel in many tasks, their assessment in Portuguese has received less attention, particularly for open-ended, discursive tasks that demand deeper reasoning and generation capabilities. While the original BLUEX benchmark addressed the scarcity of Portuguese evaluation datasets through multiple-choice questions from Brazilian university entrance exams, it did not cover the more challenging second-phase examinations, which require free-form written responses. In this work, we introduce BLUEX v2, a benchmark derived from the second-phase entrance exams of Brazil's two leading universities: UNICAMP (Comvest) and USP (Fuvest), spanning exam years 2022--2025. Our dataset comprises \textbf{395 questions} unfolding into \textbf{919 graded subquestions}, with 55.7\% of questions containing associated images (represented as context-aware captions during inference to enable evaluation across both vision-capable and text-only models). Each question is annotated with subject area, official reference answers, LLM-generated rubric criteria, and six cognitive capability tags. We evaluate 21 state-of-the-art LLMs using an LLM-as-a-judge protocol. Results reveal a 4.92-point performance spread across models (4.18--9.10 on a 0--10 scale), with Mathematical Reasoning and Image Understanding emerging as the hardest capability dimensions. The evaluation code, model outputs, and dataset are publicly available at \repourl\ and on Hugging Face at \url{https://huggingface.co/datasets/Tropic-AI/BLUEX-v2}.

\keywords{LLMs benchmark \and Portuguese \and open-ended evaluation \and large language models \and university entrance exams \and discursive questions}
\end{abstract}

\section{Introduction}
\label{sec:introduction}

The evaluation of Large Language Models (LLMs) has predominantly relied on benchmarks designed for English, leaving a significant gap in the assessment of model capabilities for other widely spoken languages. Portuguese, despite being the fifth most spoken language in the world with over 250 million native speakers, remains underrepresented in rigorous LLM evaluation~\cite{bluex2023}.

The original BLUEX benchmark~\cite{bluex2023} took an important step in addressing this gap by introducing multiple-choice questions from the first-phase entrance exams of UNICAMP and USP, Brazil's two most prestigious universities. However, the first phase tests primarily recognition and selection abilities. The \textbf{second phase}, in contrast, requires candidates to produce free-form, discursive answers demonstrating deeper understanding, multi-step reasoning, and the ability to articulate complex ideas in written Portuguese.

Second-phase exams at UNICAMP and USP present characteristics that make them particularly valuable for LLM evaluation:

\begin{itemize}
    \item \textbf{Open-ended responses}: Candidates must generate coherent structured answers, simultaneously testing understanding and generation capabilities.
    \item \textbf{Multi-step reasoning}: Questions frequently require integrating knowledge across domains, performing mathematical derivations, or constructing logical arguments over several steps.
    \item \textbf{Subject-specific depth}: Nine academic subjects are covered at the depth demanded by some of Brazil's most competitive selection processes.
    \item \textbf{Structured rubrics}: Grading criteria enable nuanced evaluation beyond binary correctness, directly grounding the automated scoring protocol.
    \item \textbf{Multimodal content}: 55.7\% of questions include figures, graphs, maps, or diagrams that are semantically essential to the answer.
\end{itemize}

Evaluating LLMs on discursive questions is more challenging than multiple-choice assessment: it requires capturing factual correctness, completeness, reasoning quality, and linguistic adequacy simultaneously. To address this, we propose an LLM-as-a-judge evaluation protocol grounded in LLM-generated rubric criteria derived from official expected answers, and empirically validated against human annotators.

This paper presents \textbf{BLUEX v2}, a benchmark of 395 discursive questions (919 subquestions) from Comvest and Fuvest exams (2022--2025), evaluated across 21 state-of-the-art models. Three empirical findings are worth highlighting up front. First, the top model (Gemini 3.1 Pro Preview) scores 9.10/10 while the weakest (LLaMA-3.2-11B Vision) scores 4.18/10, yielding a 4.92-point spread that demonstrates strong discriminative power. Second, Mathematical Reasoning (avg.\ 7.52) and Image Understanding (avg.\ 7.79) are the hardest capability dimensions, while questions with images are on average 0.54 points harder than text-only questions. Third, our LLM judge achieves 89.5\% agreement with human raters versus 94.5\% for human--human pairs, placing the automated protocol firmly in the ``substantial'' agreement range of Landis \& Koch~\cite{landis1977kappa} with \(\kappa \approx 0.69\).

Our main contributions are:

\begin{enumerate}
    \item We introduce the first open-ended benchmark derived from Brazilian university second-phase entrance exams, covering 9 subjects, 2 universities, and 4 exam years (2022--2025), including questions with multimodal content represented via context-aware image captions to enable cross-model comparability.
    \item We provide a richly annotated dataset with six cognitive capability tags per question and a four-stage construction pipeline (automated extraction, human annotation, context-aware captioning, LLM rubric generation), enabling fine-grained diagnostic analysis and full reproducibility.
    \item We propose a scalable LLM-as-a-judge evaluation protocol using LLM-generated rubric criteria derived from official expected answers, validated against two independent human reviewers, achieving substantial LLM--human agreement.
    \item We conduct a comprehensive evaluation of 21 LLMs spanning frontier and open-weight families, revealing consistent failure modes in mathematical reasoning and image understanding across model families.
    \item We quantify the multimodal penalty: image-bearing questions are 0.54 points harder on average, providing direct evidence of the multimodal challenge in a real academic setting.
\end{enumerate}

\section{Related Work}
\label{sec:related}

\subsection{Benchmarks for Portuguese and Low-Resource Languages}
\label{subsec:portuguese_benchmarks}

Standardized academic assessments have become an interesting field for gauging LLM capabilities. Notable examples include Massive Multitask Language Understanding (MMLU)~\cite{hendrycks2021mmlu}, which spans 57 tasks across Science, Technology, Engineering and Mathematics (STEM) and humanities; AGIEval~\cite{zhong2023agieval}, an aggregation of human-centric standardized tests; and GPQA~\cite{rein2024gpqa}, which targets graduate-level reasoning. These benchmarks provide a consistent metric for measuring general knowledge and logical inference in high-resource settings.

Despite the global footprint of Portuguese, its evaluation landscape remains dominated by closed-ended proxy tasks. Current assessments rely primarily on the Brazilian National High School Exam (ENEM)~\cite{nunes2023evaluating} or first-phase university entrance exams. For instance, the original BLUEX~\cite{bluex2023} and its successor, BLUEX Revisited~\cite{bluex_revisited2025}, utilize multiple-choice questions from UNICAMP and USP, with the latter incorporating image captions for basic multimodality. Although useful, these formats fail to capture the nuance of complex, generative reasoning.

\textbf{BLUEX v2 transcends these limitations by introducing a high-stakes, multi-disciplinary, and discursive benchmark.} Unlike its predecessors, it offers a multi-year, multi-subject, and multimodal evaluation based on open-ended questions. By shifting from exact-match scoring to validated automated grading of free-form text, BLUEX v2 fills a critical gap in the assessment of Portuguese-speaking LLMs, demanding a level of articulation and synthesis that multiple-choice benchmarks cannot measure.

\subsection{The Shift Toward Open-Ended Evaluation and LLM-as-a-Judge}
\label{subsec:open_ended_eval}

As models evolve beyond pattern matching, exact-match metrics have given way to nuanced, rubric-based evaluation. MT-Bench and Chatbot Arena~\cite{zheng2023judging} demonstrated the viability of LLM-as-a-judge, while G-Eval~\cite{liu2023geval} and CritiqueLLM~\cite{ke2024critiquellm} proposed scoring grounded in specific criteria. Despite their efficiency, these methods introduce validity concerns such as positional and length biases that necessitate empirical validation against human raters, as we address in Section~\ref{subsec:human_validation}.

The transition to automated grading of complex Portuguese text has recently been pioneered in specialized professional silos. The OAB Exam benchmark~\cite{pires2025oabbench} evaluates legal reasoning by focusing on the second phase of the Brazilian Bar Exam, shifting from multiple-choice to open-ended legal drafting. While successful in the legal domain, such specialized efforts do not address the multi-disciplinary reasoning required in general university entrance exams.

Validating the reliability of LLM judges in low-resource linguistic contexts remains an open research frontier. Although previous studies have focused predominantly on English, BLUEX v2 is the first to validate this protocol on Portuguese academic content through a rigorous human agreement study. By applying rubric-based judging to diverse scientific domains, we extend the scope of automated evaluation beyond specialized professional tasks to a comprehensive academic benchmark.

\section{The BLUEX v2 Benchmark}
\label{sec:dataset}

\subsection{Data Sources and Scope}
\label{subsec:data_sources}

BLUEX v2 is constructed from second-phase entrance examinations of two universities:

\begin{itemize}

\item \textbf{UNICAMP -- Comvest.} The second phase spans two days. Day~1 covers Portuguese language and literature (6 questions) plus interdisciplinary items in English and sciences. Day~2 contains 12 area-specific questions across three tracks (Biological Sciences and Health; Exact Sciences and Technology; Humanities and Arts). This information is presented in official \emph{Provas Comentadas} PDFs published by Comvest, which include both questions and the examining board's expected answers.

\item \textbf{USP -- Fuvest.} The second phase also spans two days. Day~1 covers Portuguese language and literature plus an essay. Day~2 contains subject-specific discursive questions for the student's chosen program. Questions are sourced from Fuvest's public archive; expected answers come from official \emph{Guia de Respostas Esperadas} PDFs.

\end{itemize}

The \textbf{scope} of BLUEX v2 is second-phase discursive questions from \textbf{2022 to 2025}, covering both universities. Subject classification, image captioning, and rubric generation are described in Section~\ref{subsec:pipeline}.

\subsection{Collection and Processing Pipeline}
\label{subsec:pipeline}

\begin{figure}[!htb]
    \centering
    \includegraphics[width=0.82\linewidth]{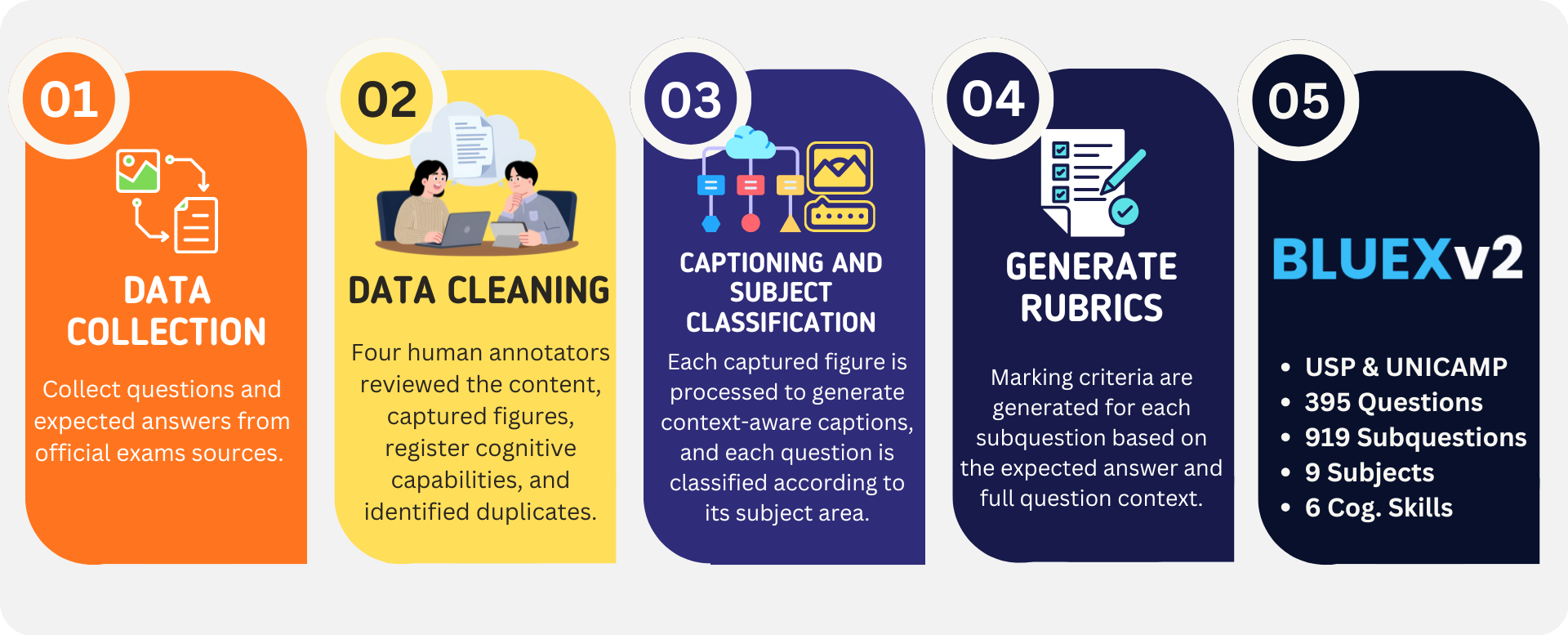}
    \caption{Dataset pipeline stages.}
    \label{fig:datasetstages}
\end{figure}

The BLUEX v2 pipeline comprises five stages, illustrated in Figure~\ref{fig:datasetstages}: Data Collection (1), Data Cleaning (2), Captioning and Subject Classification (3), Generate Rubrics (4), and the BLUEX v2 dataset release (5).

\textbf{Stage 1 -- Data Collection.}
Official exam PDFs are downloaded from the Comvest and Fuvest archives following a manifest-driven approach. Text and images are extracted using a hybrid pipeline (pdfminer \texttt{+} Azure Computer Vision OCR), questions and subquestions are segmented by regex-based heuristics, and expected answers are matched positionally across question and answer booklets. All extracted content is exported as JSON files (one per university-year).

\textbf{Stage 2 -- Data Cleaning.}
A total of 470 candidate questions were reviewed by \textbf{four independent annotators} using a custom web-based validation tool. Annotators performed: (i)~content review and correction of OCR artefacts; (ii)~verification and correction of question and subquestion segmentation; (iii)~recording of associated images per question; (iv)~assignment of six cognitive capability tags (PRK, TU, IU, MR, BK, ML); and (v)~identification and removal of duplicate questions. After deduplication and exclusion of essay (\emph{redação}), the dataset was reduced to \textbf{395 questions and 919 subquestions}.

\textbf{Stage 3 -- Captioning and Subject Classification.}
For each image, a textual description is generated using \textbf{Gemini 3.1 Flash Lite Preview} --- selected after preliminary testing as the best balance of caption quality and cost (\$0.63 for the entire dataset). The model receives the image together with the question and subquestion texts, producing \emph{context-aware captions}~\cite{bluex_revisited2025} that describe visual content in relation to the exam question rather than in isolation; these captions serve as the image modality representation during inference. Subject classification is similarly performed by querying \textbf{Sabiá-4}~\cite{laitz2026sabia4technicalreport} with the question text, yielding one of nine academic subject labels, leveraging its strong Portuguese-language understanding for Brazilian exam content.

\textbf{Stage 4 -- LLM-generated rubric criteria.}
Official expected answers provide the ground truth, but do not come with machine-readable grading rubrics. For each of the 919 subquestions, we use \textbf{Sabiá-4} to \emph{generate} a structured rubric (marking criteria) from the question text, subquestion text, and official expected answer. The model decomposes the expected answer into a list of discrete, independently checkable binary criteria. These generated rubrics are the scoring unit used by the LLM judge at evaluation time (Section~\ref{subsec:judge}).

\textbf{Stage 5 -- Dataset Release.}
The resulting BLUEX v2 dataset is publicly released at \repourl\ and as a structured Hugging Face dataset at \url{https://huggingface.co/datasets/Tropic-AI/BLUEX-v2}~\cite{santos2026bluexv2dataset}. The release includes: question JSONs with official expected answers and image captions; cognitive capability tag annotations; generated rubric criteria for all 919 subquestions; evaluation code and judge prompt iteration logs; and all 18{,}480 model outputs from the 21 evaluated models.

\subsection{Dataset Statistics}
\label{subsec:statistics}

Table~\ref{tab:dataset_stats} summarizes the composition of BLUEX v2. The annotation pipeline began with 470 candidate questions (1{,}069 candidate subquestions); after removal of duplicates and essay (\emph{redação}) by four annotators, the dataset was reduced to \textbf{395 questions totaling 919 subquestions} (avg.\ 2.33 subquestions per question), with 220 questions (55.7\%) containing at least one image.

\begin{table}[!htb]
\setlength{\tabcolsep}{1.5mm}
\renewcommand{\arraystretch}{0.92}
\centering
\caption{BLUEX v2 dataset statistics. $^\dagger$Counts non-exclusive; questions may be assigned to multiple subjects or tags. Tag acronyms: PRK=Prior Knowledge, TU=Text Understanding, IU=Image Understanding, MR=Mathematical Reasoning, BK=Brazilian Knowledge, ML=Multilingual.}
\label{tab:dataset_stats}
\footnotesize
\begin{tabular}{lrr}
\toprule
                              & \textbf{UNICAMP} & \textbf{USP} \\
\midrule
Questions                     & 211              & 184          \\
Subquestions                  & 499              & 420          \\
Questions with images         & \multicolumn{2}{r}{220 (55.7\%)} \\
Total questions               & \multicolumn{2}{r}{395} \\
Total subquestions            & \multicolumn{2}{r}{919} \\
\midrule
\multicolumn{3}{l}{\textbf{Questions by subject (combined)}$^\dagger$} \\
\multicolumn{3}{l}{\quad Sociology~(121), History~(116), Biology~(114), Geography~(102), Mathematics~(99),} \\
\multicolumn{3}{l}{\quad Physics~(75), Portuguese~(73), Chemistry~(60), Philosophy~(29)} \\
\midrule
\multicolumn{3}{l}{\textbf{Questions by cognitive capability (combined)}$^\dagger$} \\
\multicolumn{3}{l}{\quad IU~(231), PRK~(196), TU~(127), MR~(105), BK~(54), ML~(7)} \\
\bottomrule
\end{tabular}
\end{table}

\subsection{Capability Taxonomy}
\label{subsec:taxonomy}

Each question is annotated with six binary cognitive capability tags — Prior Knowledge (PRK), Text Understanding (TU), Image Understanding (IU), Mathematical Reasoning (MR), Brazilian Knowledge (BK), and Multilingual (ML) — assigned by domain experts using the validator application (Table~\ref{tab:dataset_stats}). Tags are non-exclusive. IU and PRK have the largest subquestion coverage and support robust analysis; ML (15 subquestions) is the smallest.

\section{Evaluation Methodology}
\label{sec:methodology}

\subsection{Inference Setup}
\label{subsec:inference}

Each of the 21 models was queried via API (OpenRouter / provider endpoints) for each subquestion. The input prompt is composed of: (1)~the main question text, together with the captioning content of associated images; and (2)~\emph{contextual scaffolding}: if the target subquestion is not the first item of its question (i.e., not sub-item~\textit{a}), the preceding subquestions and the model's own generated answers to them are prepended to the prompt. This mirrors the sequential structure of the original exam, where later sub-items often build on earlier ones.

\textbf{Image representation.} Rather than passing raw images directly to all evaluated models, we adopt a caption-based representation: each image is described by a context-aware textual caption generated in Stage~3 of the pipeline (Section~\ref{subsec:pipeline}). This design choice was deliberate. First, it ensures a level playing field across all 21 models, including text-only models that cannot ingest images natively. Second, context-aware captions --- produced with the full question and subquestion texts in context --- encode the semantically relevant visual content in relation to the exam question, rather than as a generic image description. Third, it preserves the multimodal character of the benchmark by keeping image-derived information in the evaluation loop for every model. We acknowledge that this approach evaluates reasoning over visual descriptions rather than raw visual perception; the implications are discussed in Section~\ref{subsec:limitations}.

Of the 919 subquestions, \textbf{39 were excluded} because their official expected answers contained images (requiring image generation, which is outside the scope of this evaluation). The remaining \textbf{880 subquestions} constitute the evaluation set, yielding $21 \times 880 = 18{,}480$ total model responses.

All models received an identical prompt template to avoid confounds (full template in the supplementary material). Models that failed to return a valid response due to API errors, timeouts, or content-filter rejections are recorded as \texttt{evaluation\_error} and assigned a score of~0 in all aggregate statistics.

\subsection{LLM-as-a-Judge Protocol}
\label{subsec:judge}

Each model response is graded by an LLM judge against the \textbf{rubric criteria} generated in Stage~4 of the pipeline (Section~\ref{subsec:pipeline}). It is important to note that these criteria are \emph{not} the official exam rubrics: they are generated by Sabiá-4 from the question, subquestion, and official expected answer, decomposing the expected answer into a list of discrete, binary-checkable items. The official expected answer therefore grounds the evaluation indirectly, through the criteria it gives rise to.

At evaluation time, the judge (Sabiá-4) receives the question context, the model's response, and the rubric criteria, and outputs a binary \texttt{is\_met} verdict (\texttt{true}/\texttt{false}) for each criterion, together with a brief justification citing the relevant passage in the model response.

Sabi\'a-4~\cite{laitz2026sabia4technicalreport} was selected as the judge for two reasons. First, it is a state-of-the-art Portuguese-native model with strong instruction-following and reasoning capabilities in Brazilian Portuguese, making it well-calibrated for evaluating responses to Brazilian university exam questions. Second, Maritaca AI provided evaluation credits that enabled running Sabi\'a-4 at full benchmark scale, which was essential for the feasibility of the evaluation pipeline.

\textbf{Scoring.} Each rubric criterion contributes an equal share to the subquestion score. For a subquestion with $k$ criteria, let $m$ be the number of criteria met. The resulting score $S = m/k \in [0, 1]$ is then scaled to a 0--10 range. The \textbf{model score} is the macro average across all 880 subquestions, with evaluation errors assigned a value of 0.

\begin{quote}
\small\textbf{Illustrative example.}
Subquestion: \emph{``Explique o papel do ATP na contração muscular.''}
Official expected answer: \emph{``O ATP fornece energia para a dissociação das pontes cruzadas entre actina e miosina.''}
Generated criteria: (C1)~\emph{ATP fornece energia para o processo}; (C2)~\emph{menciona a dissociação das pontes cruzadas}; (C3)~\emph{cita actina e miosina.}
A model response that describes ATP as an energy source and names the relevant proteins but omits the cross-bridge dissociation mechanism satisfies C1 and C3, scoring $\frac{2}{3}\times10 = 6.67$.

\end{quote}

\textbf{Threat to validity: judge-evaluatee overlap.} Sabi\'a-4 serves as rubric generator, evaluation judge, and one of the 21 evaluated models -- a potential circularity. Two factors mitigate this: criteria are grounded in \textbf{official expected answers} (not Sabi\'a-4's own outputs), and the human validation study (Section~\ref{subsec:human_validation}) shows substantial agreement (\(\kappa \approx 0.69\)) across 5 diverse model families with no evidence of systematic leniency toward any family. The choice of Sabi\'a-4 was driven by Maritaca AI's provision of evaluation credits rather than by an expectation of self-favoring scores.

The judge prompt was developed iteratively through multiple rounds of disagreement analysis against human annotations; the iteration log is available in the supplementary repository\footnote{\repourl}.

\subsection{Human Validation and Agreement Analysis}
\label{subsec:human_validation}

To validate the LLM judge as a reliable proxy for human grading, two independent reviewers (R1, R2) manually annotated the responses of \textbf{5 different models} to \textbf{11 subquestions} (covering 55 subquestions each), applying the same rubric-based binary protocol as the LLM judge. This yielded \textbf{200 criterion-level annotation pairs} used to compute pairwise agreement. We computed pairwise raw accuracy and Cohen's \(\kappa\)~\cite{cohen1960kappa} for them.

\begin{table}[!htb]
\setlength{\tabcolsep}{3.5mm}
\renewcommand{\arraystretch}{0.92}
\centering
\caption{Pairwise inter-rater agreement on the 200-rubric items validation sample. $\kappa$ follows Landis \& Koch~\cite{landis1977kappa}: 0.61--0.80 = substantial; 0.81--1.00 = almost perfect.}
\label{tab:agreement}
\footnotesize
\begin{tabular}{llcc}
\toprule
\textbf{Pair} & \textbf{Type} & \textbf{Raw Agr. (\%)} & \textbf{Cohen's} $\boldsymbol{\kappa}$ \\
\midrule
R1 vs R2              & Human--Human & 94.5 & 0.820 \\
\midrule
R1 vs Sabi\'a         & LLM--Human   & 89.5 & 0.690 \\
R2 vs Sabi\'a         & LLM--Human   & 87.0 & 0.640 \\
\bottomrule
\end{tabular}
\end{table}

Human--human agreement (\(\kappa = 0.820\)) sets the practical ceiling for the protocol. Sabi\'a reaches \(\kappa = 0.69\) vs.\ R1 --- squarely within the substantial range and comparable to validations reported for MT-Bench~\cite{zheng2023judging} on English benchmarks.

Confusion analysis reveals that the dominant judge error is a false positive (LLM marks \texttt{true} where humans mark \texttt{false}), indicating slight leniency rather than excessive strictness --- a preferable direction of bias for a benchmark aimed at ranking models.

\section{Results}
\label{sec:results}

\subsection{Overall Model Ranking}
\label{subsec:overall_results}

Table~\ref{tab:model_ranking} presents scores for the top-5 and bottom-5 models. The full ranking of all 21 models is included in \repourl.

\begin{table}[!htb]
\setlength{\tabcolsep}{4mm}
\renewcommand{\arraystretch}{0.92}
\centering
\caption{Model performance on BLUEX v2 (0--10 scale, \texttt{evaluation\_error} counted as 0). Error = per-model API error rate.}
\label{tab:model_ranking}
\footnotesize
\begin{tabular}{llcc}
\toprule
\textbf{Rank} & \textbf{Model} & \textbf{Score} & \textbf{Error (\%)} \\
\midrule
1  & Gemini 3.1 Pro Preview        & 9.10 & 0.68 \\
2  & GPT-5.4                       & 8.92 & 0.00 \\
3  & GPT-5-mini                    & 8.90 & 0.00 \\
4  & Qwen3.5-122B-A10B             & 8.83 & 0.23 \\
5  & Gemini 3.1 Flash Lite Preview & 8.83 & 0.00 \\
\midrule
17 & Microsoft phi-4 & 7.09 & 0.00 \\
18 & Qwen-2.5 7b-instruct & 6.93 & 0.00 \\
19 & cohere Command-r7b-12-2024 & 5.41 & 0.45 \\
20 & Llama-3.1-8b Instruct & 4.35 & 0.00 \\
21 & Llama-3.2-11b Vision Instruct & 4.18 & 0.00 \\
\bottomrule
\end{tabular}
\end{table}

The \textbf{performance spread} of 4.92 points (4.18--9.10) demonstrates strong discriminative power: the benchmark is neither trivially easy nor uniformly hard. Frontier API models (Gemini, GPT-5.x, Qwen3.5-122B) occupy the top tier, while smaller open-weight models (LLaMA-3.x 8--11B) cluster at the bottom, consistent with the strong correlation between model scale and language generation quality in Portuguese. The Brazilian models Sabi\'a-4 and Sabiazinho-4 achieve a performance of 8.60 and 8.57, demonstrating competitive performance despite their domain-specific focus.

\begin{figure}[!htb]
    \centering
    \includegraphics[width=0.99\linewidth]{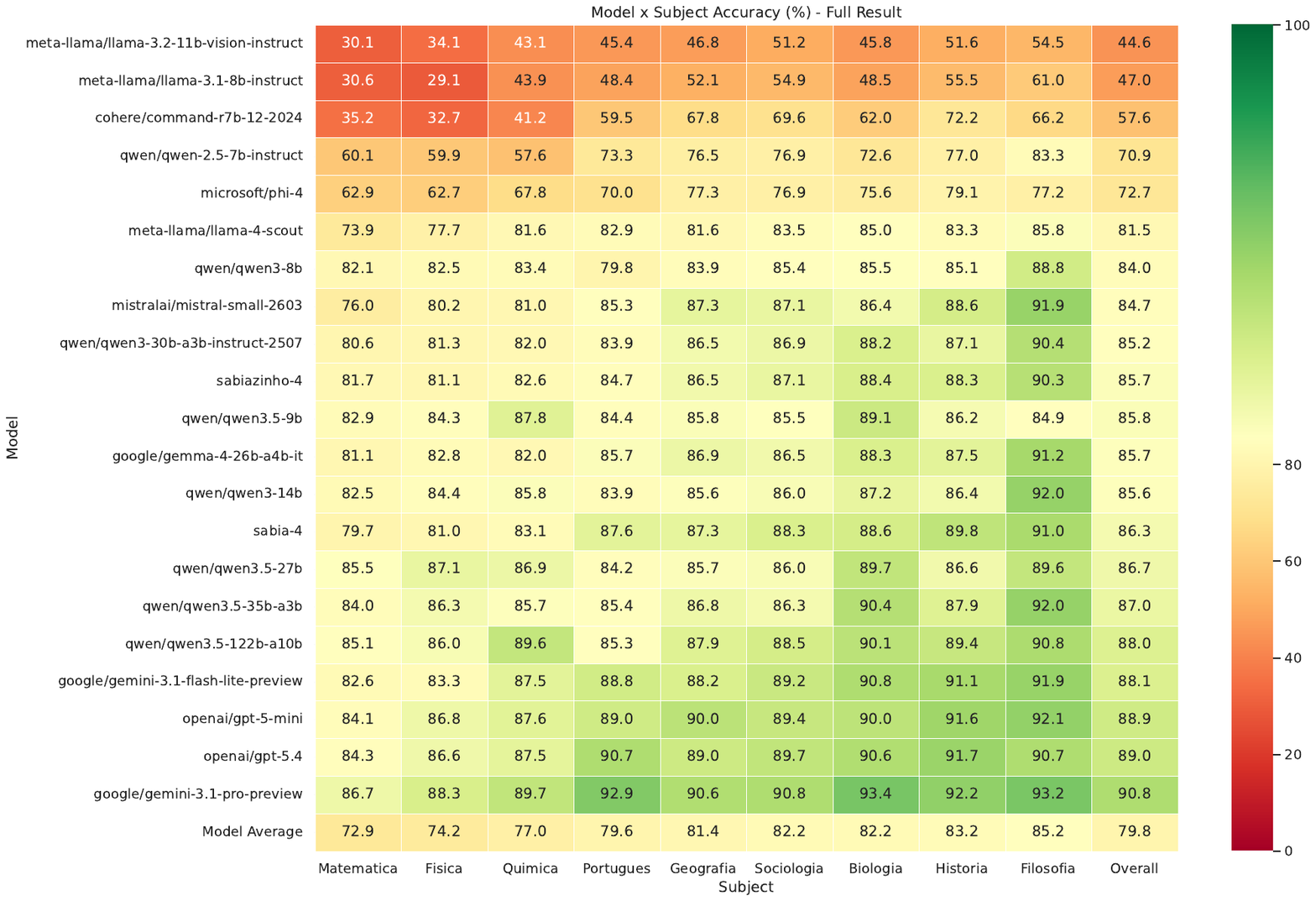}
    \caption{Performance heatmap: all 21 models (rows) $\times$ 9 subjects (columns), score 0--100. Models sorted by overall score (descending).}
    \label{fig:heatmap}
\end{figure}

Figure~\ref{fig:heatmap} shows the full performance heatmap across all 21 models and 9 subjects, illustrating both the global ranking and the per-subject variation.

\subsection{Subject-Level Difficulty}
\label{subsec:subject_results}

Table~\ref{tab:subject_difficulty} reports macro-average scores per subject (averaged over all 21 models). The 1.23-point gap between Mathematics (7.29) and Philosophy (8.52) is substantial and consistent across models, not an artifact of any single outlier. STEM subjects (Mathematics, Physics, Chemistry) cluster at the bottom, reflecting the known difficulty of symbolic reasoning and equation solving for current LLMs. Humanities subjects cluster at the top, confirming that models are better calibrated for argumentative and interpretive writing in Portuguese than for formal computation.

\begin{table}[!htb]
\setlength{\tabcolsep}{3.5mm}
\renewcommand{\arraystretch}{0.92}
\centering
\caption{Subject-level difficulty (macro-average over 21 models, 0--10 scale).}
\label{tab:subject_difficulty}
\footnotesize
\begin{tabular}{lcr}
\toprule
\textbf{Subject} & \textbf{Avg.\ Score} & \textbf{Subquestions} \\
\midrule
Mathematics   & 7.29 & 241 \\
Physics       & 7.42 & 181 \\
Chemistry     & 7.70 & 150 \\
Portuguese    & 7.96 & 139 \\
Geography     & 8.14 & 243 \\
Biology       & 8.22 & 269 \\
History       & 8.32 & 270 \\
Sociology     & 8.37 & 267 \\
Philosophy    & 8.52 &  63 \\
\bottomrule
\end{tabular}
\end{table}

\subsection{Capability-Level Analysis}
\label{subsec:capability_results}

\begin{figure}[!htb]
    \centering
    \includegraphics[width=0.96\linewidth]{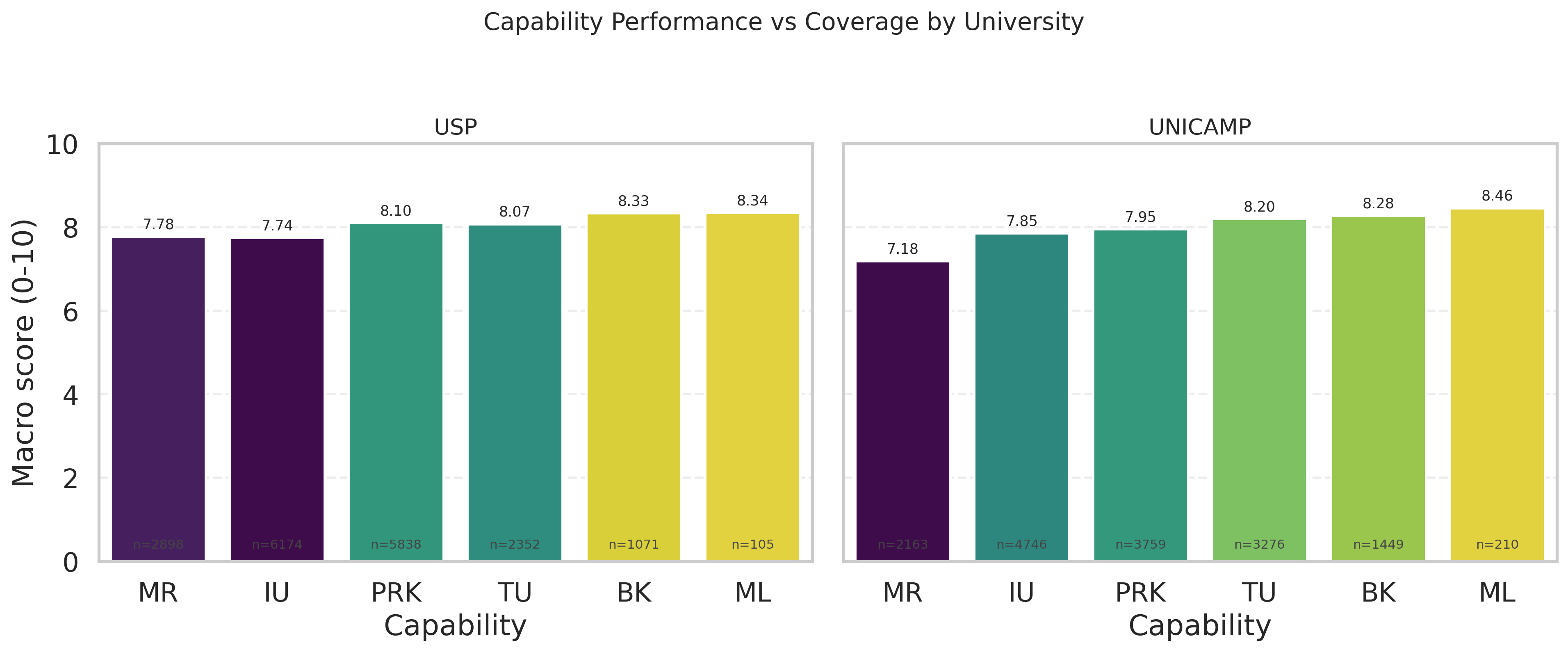}
    \caption{Capability performance broken down by university (UNICAMP vs.\ USP). The ordering of capabilities is consistent across both institutions.}
    \label{fig:capability}
\end{figure}

Table~\ref{tab:capability} and Figure~\ref{fig:capability} break down performance by the six cognitive capability tags. Three findings stand out. First, \textbf{Mathematical Reasoning (MR) is the primary model differentiator}: with a cross-model standard deviation of 1.92, MR separates top from bottom models more than any other capability --- a model scoring 9\texttt{+} overall may still perform significantly below average on MR items. Second, \textbf{Image Understanding (IU)} is the second-hardest capability (7.79), confirming that multimodal integration remains a key challenge even for frontier models. Third, \textbf{Brazilian Knowledge (BK)} scores relatively high (8.30), suggesting that recent models have absorbed sufficient Portuguese-language Brazilian content; however, this finding should be carefully interpreted given the modest sample size (122 subquestions). The ML category (15 subquestions) is the smallest.

\begin{table}[!htb]
\setlength{\tabcolsep}{4mm}
\renewcommand{\arraystretch}{0.92}
\centering
\caption{Performance by cognitive capability tag (macro-average over 21 models). Std.\ = cross-model standard deviation. $^\ddagger$Std.\ not reported for BK and ML owing to their small subquestion counts (122 and 15, respectively), where a single outlier model would dominate the variance estimate.}
\label{tab:capability}
\footnotesize
\begin{tabular}{llccr}
\toprule
\textbf{Tag} & \textbf{Name} & \textbf{Avg.\ Score} & \textbf{Std.$^\ddagger$} & \textbf{Subq.} \\
\midrule
MR  & Mathematical Reasoning & 7.52 & 1.92 & 258 \\
IU  & Image Understanding    & 7.79 & 1.45 & 552 \\
PRK & Prior Knowledge        & 8.04 & 1.40 & 476 \\
TU  & Text Understanding     & 8.14 & 1.22 & 273 \\
BK  & Brazilian Knowledge    & 8.30 & ---  & 122 \\
ML  & Multilingual           & 8.42 & ---  &  15 \\
\bottomrule
\end{tabular}
\end{table}

\subsection{Multimodal vs.\ Text-Only Performance}
\label{subsec:visual_results}

Across all models, questions \textbf{with images} receive an average score of 7.69 versus 8.23 for questions \textbf{without images} --- a \textbf{0.54-point gap}. This effect is directionally consistent across the vast majority of models and is not driven by outliers. The images in entrance exams are not decorative: they carry maps, graphs, chemical formulae, and geometric diagrams that are essential to answering correctly. Frontier vision-capable models (e.g., Gemini 3.1 Pro, GPT-5.4) exhibit a smaller image penalty, but even the best performers show non-zero degradation, corroborating the IU capability analysis in Section~\ref{subsec:capability_results}.

\subsection{Cost Analysis}
\label{subsec:cost_results}

To improve transparency and reproducibility, we report the operational cost breakdown of BLUEX v2. Dataset construction costs were modest: image captioning (Gemini 3.1 Flash Lite Preview) cost \textbf{\$0.63} and subject labeling (Sabi\'a-4) cost \textbf{\$0.23}. Table~\ref{tab:inference_costs} details per-model inference costs, totalling \textbf{\$43.00}. Judge evaluation cost approximately R\$21.00 per model ({\raise.17ex\hbox{$\scriptstyle\sim$}}\$3.95), or \textbf{\$82.95} for all 21 models. The \textbf{total pipeline cost} is \textbf{\$126.81}.

\begin{table}[!htb]
\setlength{\tabcolsep}{3.5mm}
\renewcommand{\arraystretch}{0.92}
\centering
\caption{Inference cost by model (USD), sorted by cost (descending).}
\label{tab:inference_costs}
\footnotesize
\begin{tabular}{lr@{\hspace{5mm}}lr}
\toprule
\textbf{Model} & \textbf{USD} & \textbf{Model} & \textbf{USD} \\
\midrule
Gemini 3.1 Pro Preview  & 21.41 & Sabi\'a-4         & 1.30 \\
Qwen3.5-122B            &  7.32 & Qwen3.5-9B        & 0.66 \\
Qwen3.5-27B             &  5.25 & Gemini Flash Lite & 0.55 \\
GPT-5.4                 &  4.61 & Mistral Small     & 0.34 \\
Qwen3.5-35B             &  4.08 & Sabiazinho-4      & 0.26 \\
GPT-5-mini              & 1.86  &       &  \\
\midrule
\multicolumn{3}{l}{Others (10 models)} & 1.82 \\
\multicolumn{3}{l}{\textbf{Total inference}} & \textbf{43.00} \\
\bottomrule
\end{tabular}
\end{table}

\section{Discussion}
\label{sec:discussion}

\textbf{Discriminative power and benchmark design.}
While the 4.92-point spread (4.18--9.10) distinguishes models across scales, high mean (7.93) and median (8.22) scores indicate concentration at the upper tier. This clustering suggests that while BLUEX v2 avoids a full ceiling effect, its resolution is sharper for high performers than for the lower spectrum. Nevertheless, this range exceeds typical multiple-choice Portuguese benchmarks~\cite{bluex2023}, which often saturate. This structural advantage stems from open-ended questions and partial-credit scoring, which capture nuances lost in binary formats. Finally, the capability taxonomy enhances diagnostic resolution: Mathematical Reasoning exhibits $1.6\times$ the cross-model standard deviation of Text Understanding (1.92 vs. 1.22), rank-ordering models that appear indistinguishable on aggregate scores. Performance is consistent across institutions (UNICAMP avg.\ 7.92, USP avg.\ 7.93) and stable across exam years 2022--2025, confirming the benchmark measures a stable construct that generalizes beyond any single university or cohort; pre-training contamination on earlier years cannot be fully ruled out but no performance spike is observed for 2022--2023 relative to 2024--2025.

\textbf{Why Mathematical Reasoning and Image Understanding remain hard.}
The low average scores for MR (7.52) and IU (7.79) relative to language-intensive capabilities (TU 8.14, BK 8.30) reflect a qualitative difference in the type of reasoning required. MR items in these exams demand symbolic manipulation --- algebraic derivations, geometric proofs, unit conversions --- that is inherently sequential and error-intolerant: a single algebraic slip cascades into a fully incorrect answer. This contrasts with argumentative writing, where a partially correct response can still satisfy most rubric criteria. The image penalty (0.54 points) is smaller but structurally analogous: visual understanding in these exams requires spatial interpretation (maps, diagrams, chemical structures) rather than object recognition, which frontier vision models handle considerably better. Both failure modes suggest that the benchmark will continue to differentiate models even as language generation in Portuguese improves.


\textbf{Implications for LLM development in Brazil.}
Frontier models (Gemini 3.1, GPT-5.4) perform well on Portuguese argumentation tasks, suggesting strong Portuguese pre-training. The Brazilian models Sabi\'a-4 and Sabiazinho-4 are also competitive (8.60 and 8.57) with frontier models, which speaks to the value of domain-focused Portuguese post-training. Yet MR and IU remain clear weak points across the board --- these are not uniquely Portuguese failures, but general capability gaps amplified in a high-stakes academic context where partial credit exposes them. The relatively high BK scores (8.30) indicate that Brazilian cultural and factual knowledge required in these exams is broadly accessible in large models; open-weight smaller models still lag on BK items, suggesting a knowledge-coverage rather than a reasoning deficit for that category.

\subsection{Limitations}
\label{subsec:limitations}

\begin{itemize}

\item \textbf{LLM rubric generation and judge overlap.} Rubric criteria are generated by Sabi\'a-4 from official expected answers and evaluated by the same model acting as judge --- which is also one of the evaluated models. Although criteria are grounded in official answers and human validation shows no cross-family bias (Section~\ref{subsec:human_validation}), the 5--8 pp LLM--human agreement gap implies roughly 10--13\% of borderline verdicts may be noisy, limiting fine-grained per-subquestion analysis.

\item \textbf{Caption-based multimodality.} Because evaluated models receive context-aware textual captions rather than raw images (Section~\ref{subsec:inference}), the benchmark measures reasoning over visual descriptions rather than direct visual perception. This design prioritises cross-model comparability but means that models with strong native vision encoders may be underestimated on IU items. Additionally, caption quality is tied to Gemini 3.1 Flash Lite: if captions omit or distort visual details, IU performance may be further underestimated. Future versions should include a raw-image evaluation track to disentangle caption quality from model vision capability.

\item \textbf{Benchmark saturation risk.} Top scores (currently 9.10) may approach the ceiling as models improve. A harder adversarial subset should be considered in future iterations.

\end{itemize}
\section{Conclusions}
\label{sec:conclusion}

We introduced BLUEX v2, the first multimodal, open-ended benchmark for evaluating LLMs on discursive questions from Brazilian university second-phase entrance exams, covering 9 subjects, 2 universities, and 4 exam years 2022--2025 (Contribution~1). The dataset comprises 395 questions (919 subquestions) annotated with official expected answers, LLM-generated rubric criteria, and six cognitive capability tags, enabling fine-grained diagnostic analysis across languages, subjects, and capabilities (Contribution~2). An LLM-as-a-judge protocol grounded in these rubric criteria was validated against two human reviewers across 200 criterion-level comparisons, achieving substantial agreement with 89.5\% LLM--human vs. 94.5\% human--human, confirming the protocol's reliability for scalable, annotation-free evaluation (Contribution~3).

Evaluating 21 state-of-the-art models (Contribution~4) reveals that Mathematical Reasoning (avg.\ 7.52) and Image Understanding (avg.\ 7.79) are the hardest capability dimensions and the strongest differentiators, while a 4.92-point performance spread confirms the benchmark discriminates effectively without ceiling or floor effects. Questions containing images are 0.54 points harder on average across all models (Contribution~5), quantifying the real-world cost of multimodal reasoning in a high-stakes academic context.

We release the dataset, evaluation code, judge prompt, and model outputs to foster further research on Portuguese language understanding and generation, as well as to promote transparency and reproducibility\footnote{\repourl\ --- \url{https://huggingface.co/datasets/Tropic-AI/BLUEX-v2}}.

Directions for future work include extending the dataset to 2018--2021 exam years, developing a live leaderboard for continuous model submissions, and investigating prompt sensitivity and adversarial perturbations.

\begin{credits}
\subsubsection{\ackname} We thank Maritaca AI for providing the evaluation credits that enabled running Sabiá-4 as the production judge at full benchmark scale and Instituto Kunumi for their partial support.

\subsubsection{\discintname}
Some of the authors declare that they are affiliated with Maritaca AI, the company responsible for Sabia-4 and Sabiazinho-4, both of which were evaluated in this work.
\end{credits}

\bibliographystyle{splncs04}
\bibliography{references}

@inproceedings{bluex2023,
  author    = {Thales Sales Almeida and Thiago Laitz and Giovana K. Bon\'{a}s and Rodrigo Nogueira},
  title     = {{BLUEX: A Benchmark Based on {B}razilian Leading Universities Entrance Exams}},
  booktitle = {Intelligent Systems (BRACIS 2023)},
  series    = {Lecture Notes in Computer Science},
  volume    = {14195},
  pages     = {337--347},
  publisher = {Springer},
  year      = {2023},
  doi       = {10.1007/978-3-031-45368-7_22}
}

@inproceedings{bluex_revisited2025,
  author    = {Joao Guilherme Alves Santos and Giovana Kerche Bon\'{a}s and Thales Sales Almeida},
  title     = {{BLUEX Revisited: Enhancing Benchmark Coverage with Automatic Captioning}},
  booktitle = {Proceedings of ENIAC},
  year      = {2025},
  eprint    = {2508.21294},
  archivePrefix = {arXiv},
  primaryClass  = {cs.CL}
}

@inproceedings{hendrycks2021mmlu,
  author    = {Dan Hendrycks and Collin Burns and Steven Basart and Andy Zou and Mantas Mazeika and Dawn Song and Jacob Steinhardt},
  title     = {{Measuring Massive Multitask Language Understanding}},
  booktitle = {International Conference on Learning Representations (ICLR)},
  year      = {2021}
}

@article{zhong2023agieval,
  author    = {Wanjun Zhong and Ruixiang Cui and Yiduo Guo and Yaobo Liang and Shuai Lu and Yanlin Wang and Amin Saied and Weizhu Chen and Nan Duan},
  title     = {{AGIEval: A Human-Centric Benchmark for Evaluating Foundation Models}},
  journal   = {arXiv preprint arXiv:2304.06364},
  year      = {2023}
}

@article{rein2024gpqa,
  author    = {David Rein and Betty Li Hou and Asa Cooper Stickland and Jackson Petty and Richard Yuanzhe Pang and Julien Dirani and Julian Michael and Samuel R. Bowman},
  title     = {{GPQA: A Graduate-Level Google-Proof Q\&A Benchmark}},
  journal   = {arXiv preprint arXiv:2311.12022},
  year      = {2024}
}

@article{nunes2023evaluating,
  author    = {Desnes Nunes and Ricardo Primi and Ramon Pires and Roberto Lotufo and Rodrigo Nogueira},
  title     = {{Evaluating GPT-3.5 and GPT-4 Models on Brazilian University Admission Exams}},
  journal   = {arXiv preprint arXiv:2303.17003},
  year      = {2023}
}

@inproceedings{zheng2023judging,
  author    = {Lianmin Zheng and Wei-Lin Chiang and Ying Sheng and Siyuan Zhuang and Zhanghao Wu and Yonghao Zhuang and Zi Lin and Zhuohan Li and Dacheng Li and Eric P. Xing and Hao Zhang and Joseph E. Gonzalez and Ion Stoica},
  title     = {{Judging LLM-as-a-Judge with MT-Bench and Chatbot Arena}},
  booktitle = {Advances in Neural Information Processing Systems (NeurIPS)},
  year      = {2023}
}

@article{pires2025oabbench,
      title={{Automatic Legal Writing Evaluation of LLMs}}, 
      author={Ramon Pires and Roseval Malaquias Junior and Rodrigo Nogueira},
      year={2025},
      journal={arXiv preprint arXiv:2504.21202},
}

@article{landis1977kappa,
  title={{An Application of Hierarchical kappa-type Statistics in the Assessment of Majority Agreement among Multiple Observers}},
  author={Landis, J Richard and Koch, Gary G},
  journal={Biometrics},
  pages={363--374},
  year={1977},
  publisher={JSTOR}
}

@article{cohen1960kappa,
  title={{A Coefficient of Agreement for Nominal Scales}},
  author={Jacob Cohen},
  journal={Educational and Psychological Measurement},
  year={1960},
  volume={20},
  pages={37 - 46},
}

@misc{liu2023geval,
      title={{G-Eval: NLG Evaluation using GPT-4 with Better Human Alignment}}, 
      author={Yang Liu and Dan Iter and Yichong Xu and Shuohang Wang and Ruochen Xu and Chenguang Zhu},
      year={2023},
      eprint={2303.16634},
      archivePrefix={arXiv},
      primaryClass={cs.CL},
      url={https://arxiv.org/abs/2303.16634}, 
}

@article{ke2024critiquellm,
      title={{CritiqueLLM: Towards an Informative Critique Generation Model for Evaluation of Large Language Model Generation}}, 
      author={Pei Ke and Bosi Wen and Zhuoer Feng and Xiao Liu and Xuanyu Lei and Jiale Cheng and Shengyuan Wang and Aohan Zeng and Yuxiao Dong and Hongning Wang and Jie Tang and Minlie Huang},
      year={2024},
      journal={arXiv preprint arXiv:2311.18702},
}

@article{laitz2026sabia4technicalreport,
      title={Sabi\'a-4 Technical Report}, 
      author={Thiago Laitz and Thales Sales Almeida and Hugo Abonizio and Roseval Malaquias Junior and Giovana Kerche Bonás and Marcos Piau and Celio Larcher and Ramon Pires and Rodrigo Nogueira},
      year={2026},
      journal={arXiv preprint arXiv:2603.10213},
}

@misc{santos2026bluexv2dataset,
  author       = {Jo\~ao Guilherme Alves Santos and Giovana Kerche Bon\'as and Thiago Laitz and Thales Sales Almeida and Helio Pedrini},
  title        = {{BLUEX-v2}: Benchmarking {LLMs} on Open-Ended Questions from {Brazilian} University Entrance Exams},
  year         = {2026},
  publisher    = {Hugging Face},
  howpublished = {\url{https://huggingface.co/datasets/Tropic-AI/BLUEX-v2}},
}

\end{document}